\documentclass{article}

\usepackage{spconf,amsmath,graphicx}
\usepackage[margin=1in]{geometry} 
\usepackage{comment} 
\usepackage{longtable}
\usepackage{array}
\usepackage{multirow}
\usepackage{booktabs} 
\usepackage{makecell} 
\usepackage{algorithm}
\usepackage{algorithmic}
\usepackage{subcaption} 
\usepackage{hyperref}

\usepackage{spconf}
\title{Image-Difficulty-Aware Evaluation of Super-resolution Models}

\name{Author(s) Name(s)\thanks{}}
\address{Author Affiliation(s)}
 \name{Atakan Topaloğlu$^*$ \qquad Ahmet Bilican$^*$ \qquad Cansu Korkmaz \qquad A. Murat Tekalp \thanks{$^*$ Equal contribution. \\ This work was supported in part by an AI Fellowship to C. Korkmaz provided by the KUIS AI Center. A.M. Tekalp acknowledges support from Turkish Academy of Sciences (TUBA).}}
 \address{College of Engineering and KUIS AI Center, Koç University, Istanbul, Turkey}

\begin{document}
\maketitle
\thispagestyle{empty}

\begin{abstract}
Image super-resolution models are commonly evaluated by average scores (over some benchmark test sets), which fail to reflect the performance of these models on images of varying difficulty and that some models generate artifacts on certain difficult images, which is not reflected by the average scores. We propose difficulty-aware performance evaluation procedures to better differentiate between SISR models that produce visually different results on some images but yield close average performance scores over the entire test set. In particular, we propose two image-difficulty measures, the high-frequency index and rotation-invariant edge index, to predict those test images, where a model would yield significantly better visual results over another model, and an evaluation method where these visual differences are reflected on objective measures. Experimental results demonstrate the effectiveness of the proposed image-difficulty measures and evaluation methodology.
\end{abstract}


\section{Introduction} \vspace{-4pt}
\label{sec:intro}
Proper evaluation of image super-resolution (SR) models is a challenging problem. Image SR models are typically trained on popular benchmark training sets such as DIV2K \cite{Agustsson_2017_CVPR_Workshops} and/or LSDIR \cite{cite_LSDIR}. Then, they are evaluated by means of fidelity and/or perceptual measures averaged over the corresponding test sets. Images contain various structured high spatial-frequency content, such as edges and textures, which determine the difficulty level of super resolving them. Yet, commonly used training and evaluation procedures rarely take the difficulty of individual images into account. As a result, we observe that  some models yield noticeable visual artifacts on certain difficult images, but the~average PSNR values reported over the entire test set fail to reflect the visual quality differences observed on these difficult images.

\begin{figure}
    \centering
    \begin{subfigure}{0.24\columnwidth}
        \includegraphics[width=\columnwidth]{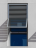}
        \caption{}
    \end{subfigure}
    \begin{subfigure}{0.24\columnwidth}
        \includegraphics[width=\columnwidth]{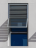}
        \caption{}
    \end{subfigure}
    \begin{subfigure}{0.24\columnwidth}
        \includegraphics[width=\columnwidth]{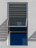}
        \caption{}
    \end{subfigure}
    \begin{subfigure}{0.24\columnwidth}
        \includegraphics[width=\columnwidth]{figures/window_zoom/0000065x4_test_WaveletFormer_X4_bigg_image_mixed_zoom.pdf}
        \caption{}
    \end{subfigure} \vspace{-10pt}
    \caption{
    Demonstration of the fact that the average PSNR over validation/test set does not reveal the performance difference between two models using a crop from image~65 from LSDIR \cite{cite_LSDIR} validation set: (a)~high-resolution ground-truth crop, (b) super resolved crop using the edge model (average PSNR 26.1842 dB), (c) super resolved crop using the global model (average PSNR 26.1825~dB).
    }
    \label{fig:edge_vs_mixed}
\end{figure}

We illustrate this problem in Figure \ref{fig:edge_vs_mixed}, where we compare the performances of two SR models, a global model trained over the entire dataset and a special model trained over edge patches (selected by using the measures introduced in Section~\ref{sec:method_difficulty_measure}).
We can see that the special model yields significantly better visual results with much less artifacts compared to the global model, while the average PSNR of the two models differ only in the thousandths digit. Looking at the average PSNR values, one may erroneously conclude that there is no significant difference between the performances of these two models. 

In order to provide more reliable evaluation of SR models, Section~\ref{sec:method_difficulty_measure} proposes two image-difficulty measures, the~high-frequency index and rotation-invariant edge index, to predict test images, where a model would yield better results over another. Based on these measures and top 1\% MSE criterion, in Section~\ref{sec:diff_aware_perf_eval}, we introduce an image-difficulty-aware SR evaluation method, where visual differences in difficult images are reflected on objective measures. Experimental results in Section~\ref{sec:results} demonstrate the effectiveness of the proposed image-difficulty measures and evaluation methodology.  \vspace{-7pt}

\section{Related Works} \vspace{-4pt}
\label{sec:related_works}

We can broadly classify SR models as regressive or generative based on convolutional and/or transformer architectures. While none of these models can fully recover high-frequency details and all models generate artifacts, it is well-known that regressive models generate images with blurry high-frequency (HF) details and generative models are prone to yielding HF hallucinations. 

Convolutional SR architectures have been improved by using
U-Net, residual/dense connections~\cite{tong_densenet, zhang_res_dense}, and attention mechanisms~\cite{zhang2018rcan, niu_han}, to boost fidelity and perceptual quality. Transformer-based architectures, such as SwinIR~\cite{Liang2021SwinIRIR} and HAT~\cite{chen2023activating}, improve performance by modeling long-range dependencies. Artifact mitigation methods include ESRGAN+ \cite{ESRGAN_plus} that exploits stochastic variation by injecting noise to inputs, LDL~\cite{details_or_artifacts} that predict artifact probabilities from patch-level residuals and SROOE~\cite{srooe_Park_2023_CVPR} that optimize networks with perceptual and objective image maps. Fourier-domain approaches, including ESRGAN-FS~\cite{freq_sep} and CARB GAN-FS~\cite{GuidedFreqSep}, address specific frequency components but cannot exploit spatial localization of artifacts. Wavelet-domain methods, such as PDASR~\cite{PDASR} and methods using the wavelet loss, such as WGSR~\cite{WGSR}, aim to reconstruct genuine image details while suppressing HF artifacts.

Recent advancements in SR evaluation include \cite{cheng2023newsuperresolutionmeasurementperceptual, srqc, Wu_2024_CVPR}. Despite their sophistication, these methods often culminate in a single performance figure per dataset. This practice, common in SR evaluation, tends to obscure how model performance varies significantly with image difficulty and characteristics, thereby failing to capture critical nuances that a difficulty-aware evaluation, as proposed herein, can reveal.

SR models are typically evaluated by fidelity metrics, such as PSNR and SSIM, or perceptually-oriented metrics, such as LPIPS~\cite{lpips} and NIQE~\cite{niqe}. These metrics, typically averaged over benchmark datasets, are widely used to quantify pixel-level accuracy or perceptual quality. However, evaluation based on average scores fails to account for variations in image difficulty across datasets~\cite{9897278}; hence, does not reflect visually annoying artifacts on challenging images. This underscores the need for difficulty-aware evaluation to better assess model performance across varying image complexities. \vspace{-4pt}

\section{Image Difficulty Measures} \vspace{-4pt}
\label{sec:method_difficulty_measure}

We propose the High-Frequency Index (HFI) to determine images with significant HF content and the~Rotation-Invariant Edge Index (RIEI) to further classify images with HF content as edge vs. texture images in order to measure the~difficulty level of super resolving an image using only the observed LR image.  \vspace{-7pt}

\subsection{High-Frequency Index (HFI)} 
\begin{figure}[t]
    \centering
    \includegraphics[width=1\linewidth]{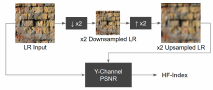}
    \vspace{-6pt}
    \caption{HF-Index Computation}
    \label{fig:HFI_computation}
\end{figure}

\vspace{-3pt}
\quad  HFI is a measure of high-frequency content of the~observed LR image. It is based on the assumption that the high-frequency content of the HR image can be estimated from the difference between the observed LR image and the image obtained by bilinear interpolation of $\times$2 down-sampled version of the LR image. The smaller this difference (measured by PSNR in dB), the lower the~high-frequency content of the HR image; hence the~easier to super resolve the image. The block diagram illustrating computation of HFI is depicted in Figure \ref{fig:HFI_computation}. Our claim that HFI is an indicator of SR-difficulty is empirically supported by the~SR-PSNR vs. HFI plot depicted in Figure \ref{fig:psnrlrpsnrcorr}, where the Pearson correlation between SR-PSNR and HFI is 0.665 and the Spearman correlation is 0.614.

\begin{figure}[t]
    \centering
        \includegraphics[width=\linewidth]{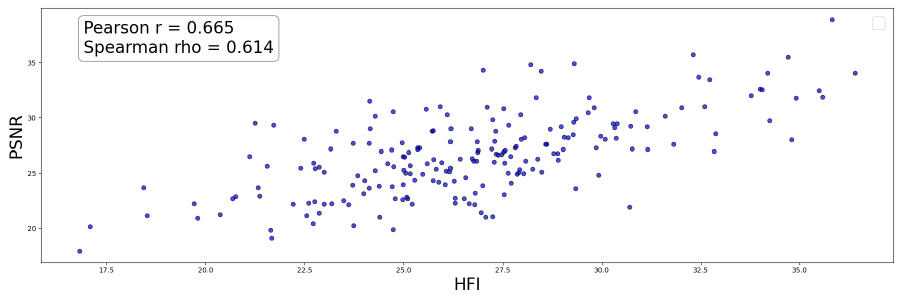} 
        \label{fig:psnrlrpsnrcorr} \vspace{-22pt}
    \caption{SR-PSNR vs HFI for $\times$4 super resolving images in the LSDIR validation set using SwinIR \cite{Liang2021SwinIRIR}.}
    \label{fig:psnrlrpsnrcorr}
\end{figure}

\vspace{-3pt}

\begin{figure}[ht]
    \centering
    \includegraphics[width=0.87\linewidth]{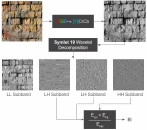}
    \label{fig:alpha_computation} \vspace{-5pt}
    \caption{Edge Index (EI) Computation}
    \label{fig:alpha_computation}
\end{figure}

\vspace{-10pt}
\subsection{Rotation-Invariant Structured Edge Index (RIEI)}

The dominant characteristic of images containing significant HF content may be structured edges, such as text and delineations of buildings, windows, or other architectural demarcations, or texture patterns, such as hair, animal fur, and others. The former are prone to ringing artifacts and color distortions, whereas latter tend to manifest blurring in the SR results. We propose RIEI to differentiate between these two types of high-frequency content using only the low-resolution (LR) image. 

We first introduce edge index (EI) based on the wavelet subbands. To this effect, we compute the wavelet transform of the Y-channel of the observed LR image, as illustrated in Figure~\ref{fig:alpha_computation}, to generate the LL, LH, HL and HH subbands, where HL and LH subbands represent edge-like oriented high frequencies, while HH subband corresponds to texture-like HF features. We have chosen the Symlet-19 wavelet following Korkmaz {\it et al.} \cite{WGSR}, who demonstrated 
that the Symlet 19 wavelet represents local HF features effectively. The EI is defined by \vspace{-3pt}
\begin{equation}
EI=\frac{E_{LH}+E_{HL}}{E_{HH}}
\end{equation}

where $E_{\cdot}$ denotes the sum of absolute values of the coefficients in the respective subband. Higher and lower EI values indicate edge and texture features, respectively.

We observed that EI is not invariant to the orientation of dominant edges in the image.
To overcome this limitation, we introduce the Rotationally Invariant Edge Index (RIEI), whereby the LR image is rotated in increments of 20° (from 0° to 80°), and the maximum EI value across these rotations is selected. \vspace{-5pt}
\vspace{-3pt}

\begin{equation}
\text{RIEI} = max(\text{EI}_\theta)
\end{equation}

\vspace{-6pt}

\begin{figure}[b!]
    \centering
    \begin{subfigure}{0.45\columnwidth}
        \includegraphics[width=\columnwidth]{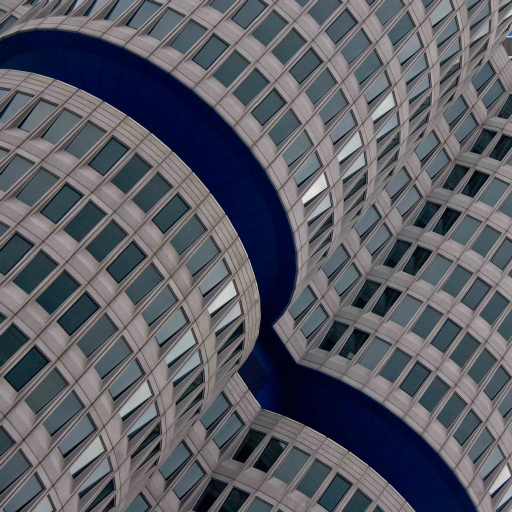}
        \caption{}
    \end{subfigure}
    \begin{subfigure}{0.45\columnwidth}
        \includegraphics[width=\columnwidth]{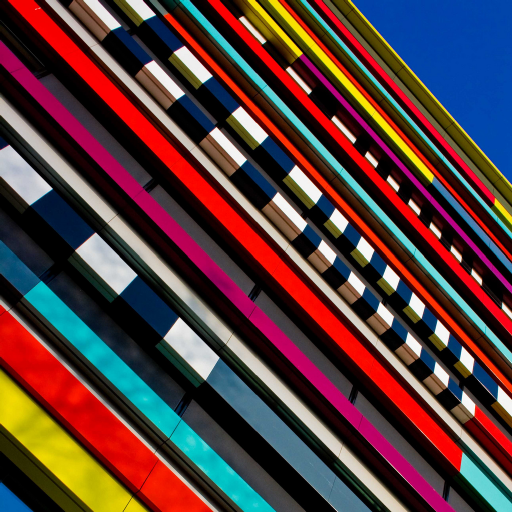}
        \caption{}
    \end{subfigure}
    \caption{Images with 45$^\circ$ edges from Urban 100 Dataset where RIEI gives reliable scores. (a) Image 068 (RIEI: 6.240,  EI: 1.743) (b) Image 081 (RIEI: 7.943, EI: 1.311)}
    \label{fig:riei_fail}
\end{figure}

\begin{figure}
    \centering
    \begin{subfigure}{0.45\columnwidth}
        \includegraphics[width=\columnwidth]{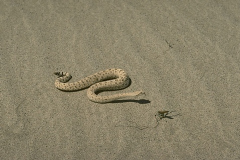}
        \caption{}
    \end{subfigure}
    \begin{subfigure}{0.45\columnwidth}
        \includegraphics[width=\columnwidth]{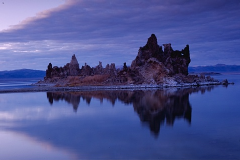}
        \caption{}
    \end{subfigure}

        \begin{subfigure}{0.45\columnwidth}
        \includegraphics[width=\columnwidth]{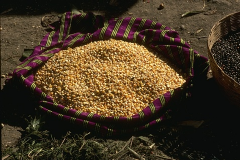}
        \caption{}
    \end{subfigure}
    \begin{subfigure}{0.45\columnwidth}
        \includegraphics[width=\columnwidth]{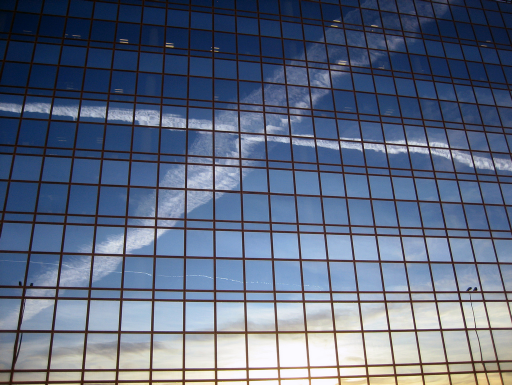}
        \caption{}
    \end{subfigure}
    \caption{Examples of images classified by their HFI vs. RIEI values. (a) easy-texture  BSD100 037.pdf (b) easy-edge BSD100 \cite{bsd100_cite} 018.pdf (c) hard-texture  BSD100 084.pdf  (d) hard-edge Urban100 \cite{urban100_cite} 055.pdf}
    \label{fig:riei_lrpsnr_4_images}
\end{figure}

\noindent The RIEI captures edges more robustly than EI and independent of orientation as shown in Figure \ref{fig:riei_fail}, where we see that edges with 45$^\circ$ orientation yield small EI values.

\begin{table}[b!]
\centering
\begin{tabular}{l|cc|cc}
\hline
\textbf{Test Dataset}  & \textbf{HFI} & \textbf{HFI} & \textbf{RIEI} & \textbf{RIEI} \\
& mean & med & mean & med  \\ \hline 
manga109            & 25.458 & 25.178 & 6.050 & 5.985 \\
urban100            & 24.668 & 24.886 & 5.971 & 5.452 \\
lsdir\_valid        & 26.288 & 25.958 & 5.669 & 5.424 \\
div2k\_test         & 28.570 & 28.115 & 5.430 & 5.291 \\
div2k\_valid        & 28.429 & 28.042 & 5.419 & 5.303 \\
bsd100              & 28.810 & 28.044 & 5.368 & 5.215 \\ \hline
\end{tabular} 
\caption{Mean and median HFI and RIEI values.}
\label{tab:mean_stats}
\end{table}

In order to show that RIEI can distinguish between textures and edges, we manually extracted and labeled 1242 patches (512$\times$512) with edge or texture content from images in the LSDIR \cite{cite_LSDIR} training set. The~median RIEI to separate edge and textures was 5.14. On a hold-out set of 292 patches, the classification accuracy was 83\%.
Figure~\ref{fig:riei_lrpsnr_4_images} shows examples of each image category, characterized in terms their RIEI and HFI values.
The results in Table \ref{tab:mean_stats} indicate a clear relationship between dataset characteristics and the proposed metrics. Notably, datasets characterized by pronounced high-frequency edge content, such as Manga109 and Urban100, demonstrate elevated RIEI scores. 
Moreover, these datasets exhibit comparatively lower HFI, suggesting that the inherent complexity of their visual content poses significant challenges for super-resolution models,
which is affirmed by the lower PSNR values models attain 
on these datasets.
Conversely, datasets such as BSD100 and DIV2K (both test and validation) report higher HFI values, indicating 
they contain less abrupt transitions,
making them more amenable to effective recovery by super-resolution methods.
\vspace{-5pt}

\vspace{-5pt}
\section{Difficulty-Aware Model Evaluation}
\label{sec:diff_aware_perf_eval}
\vspace{-5pt}
\quad The average PSNR measure over a test set treats every image in the test set and every pixel in every image as equals and assigns them equal weight. We propose a methodology to evaluate SR models based on selected images (by their global properties) as well as on selected pixels of images (by their sparse local properties) for a more complete assessment of SR models. To this effect, we propose quadrant-based evaluation in the HFI vs. RIEI plane for selective analysis based on global image properties, and PSNR99 measure for selective analysis based on sparse local image properties.

\begin{figure}
    \centering
    \begin{subfigure}{0.48\columnwidth}
        \includegraphics[width=\columnwidth]{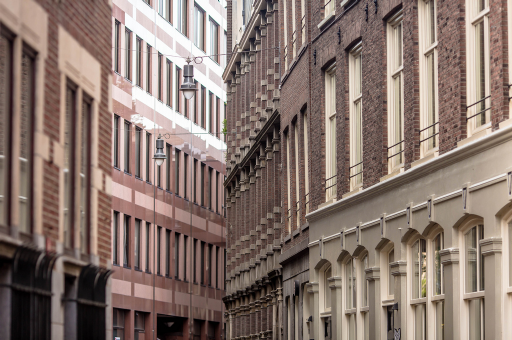}
        \caption{}
    \end{subfigure}
    \begin{subfigure}{0.425\columnwidth}
        \includegraphics[width=\columnwidth]{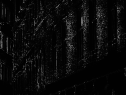}
        \caption{}
    \end{subfigure}
    \begin{subfigure}{0.48\columnwidth}
        \includegraphics[width=\columnwidth]{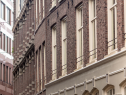}
        \caption{}
    \end{subfigure}
    \begin{subfigure}{0.48\columnwidth}
        \includegraphics[width=\columnwidth]{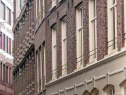}
        \caption{}
    \end{subfigure}
    \caption{Artifact map based on PSNR99 accurately captures visually disturbing areas of the SR image (e.g., hallucinations on the bricks). (a) HR Image (b) Zoomed in PSNR99 error map (c) Zoomed in HR crop (d) Zoomed in SR crop (PSNR = 21.19 dB PSNR99 = 8.38 dB)} \vspace{-10pt}
    \label{fig:image64}
\end{figure}
 \vspace{-12pt}
\subsection{HFI vs RIEI Quadrant-Based Evaluation} \vspace{-8pt}
To perform difficulty-aware evaluation of SR models, we characterize images based on their RIEI and HFI, where higher RIEI indicates edge-dominance and lower HFI (in dB) implies higher estimated difficulty of super resolution.
We then group images based on their HFI and RIEI values into four quadrants separated by the medians of HFI and RIEI, respectively. Analyzing the mean PSNR of images in each quadrant provides us information on how an SR model performs on images in each content different category. Results are shown in Section~\ref{sec:results}.
\vspace{-12pt}
\subsection{Sparse Local Artifact Evaluation: PSNR99} 
\vspace{-8pt}The artifacts that are most annoying to the human eye are sparse and localized. Evaluation based on full images often dilutes the impact of localized artifacts that are perceptually significant, such as ringing around sharp edges, blocking artifacts, or GAN-induced hallucinations. To address this problem, we propose PSNR99, a metric that focuses on the largest errors effectively capturing those regions of images that are most visually disturbing and often overlooked in conventional evaluations. 
We compute PSNR99 on the Y channel of images by rank-ordering pixel-wise squared errors and taking the average of top 1$\%$ of them as shown in Algorithm~\ref{algorithm1}.
Figure~\ref{fig:image64} demonstrates the ability of PSNR99 to localize and quantify visually disturbing artifacts of ESRGAN+ \cite{ESRGAN_plus} on Urban100 \cite{urban100_cite} test set image 64. 
\vspace{-14pt}
\section{Experimental Results}
\label{sec:results}
\vspace{-8pt}

We present three case studies using the methodology proposed in Section~\ref{sec:diff_aware_perf_eval} to demonstrate inadequacy of average measures over test sets. They are: 1. Analysis of PSNR results for ESRGAN+; 2. Comparison of pixel-loss vs. wavelet loss; 3. Comparison of GAN vs. diffusion models, taking image characteristics into consideration.
\vspace{8pt}

\begin{figure}[b!]
\centering
        \includegraphics[width=1.05\columnwidth]{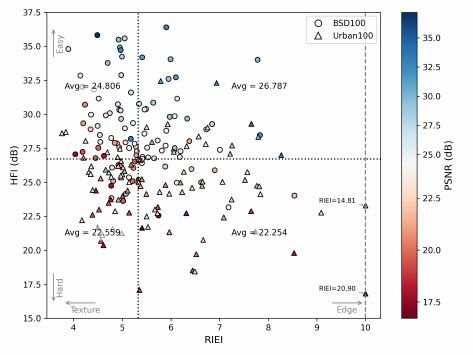} \vspace{-20pt}
        \caption{Quadrant-based PSNR analysis of ESRGAN+ results via HFI vs. RIEI scatter plot, where PSNR values are color-coded, on combined BSD100 and Urban100.}
        \label{fig:only_esrgan_psnr}
\end{figure} \vspace{-3pt}
\begin{table}[b!]
\centering
\resizebox{\columnwidth}{!}{
\begin{tabular}{lccccc}
\toprule
\multicolumn{1}{c}{} & \multicolumn{2}{c}{\textbf{Easy}} & \multicolumn{2}{c}{\textbf{Hard}} & \multirow{2}{*}{\makecell{\textbf{Global}\\\textbf{Average}}} \\
\cmidrule(lr){2-3}\cmidrule(lr){4-5}
\multicolumn{1}{c}{} & \textbf{Texture} & \textbf{Edge} & \textbf{Texture} & \textbf{Edge} &  \\
\midrule
PSNR     & 24.806  & 26.787  & 22.559   & 22.254  &  24.079   \\
PSNR99   & 11.619 & 12.617  & 9.040  & 8.978   &  10.553\\
\bottomrule
\end{tabular} }\vspace{-5pt}
\caption{Summary of quadrant-based PSNR analysis of ESRGAN+ results on Urban100 and BSD100.
}
\label{tab:esgran_clipiqa}
\end{table}

\vspace{-3pt}
\noindent \underline{Case 1:} We analyze PSNR performance of ESRGAN+ as a function of HFI and RIEI in Figure~\ref{fig:only_esrgan_psnr} and Table~\ref{tab:esgran_clipiqa}. Inspection of Figure~\ref{fig:only_esrgan_psnr} shows that the PSNR values for individual images in  BSD100 and Urban100 test sets show a large variation. Hence, instead of single average PSNR for all images, we report quadrant-based average PSNR values on the HFI vs. RIEI plane, which is a compromise between a single average for all images vs. reporting individual PSNR for each image. The quadrants are determined by the median HFI and RIEI values indicated by dotted lines in Figure \ref{fig:only_esrgan_psnr} and labeled as Easy-Texture, Hard-Texture, Easy-Edge and Hard-Edge as in Table~\ref{tab:esgran_clipiqa}. We~observe in Table~\ref{tab:esgran_clipiqa} that while the PSNR for Easy-Texture class is close to the global average, the PSNR average for the other three quadrants differ from the global average by about 2.5 dB.

\begin{figure}
    \centering
    \begin{subfigure}{0.48\columnwidth}
        \includegraphics[width=\columnwidth]{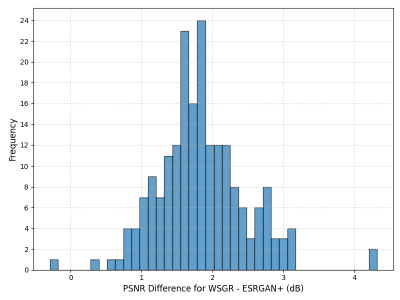}
        \caption{}
    \end{subfigure}
    \begin{subfigure}{0.48\columnwidth}
        \includegraphics[width=\columnwidth]{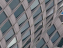}
        \caption{}
    \end{subfigure}
    \begin{subfigure}{0.48\columnwidth}
        \includegraphics[width=\columnwidth]{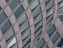}
        \caption{}
    \end{subfigure}
    \begin{subfigure}{0.48\columnwidth}
        \includegraphics[width=\columnwidth]{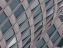}
        \caption{}
    \end{subfigure}
    \caption{Per-image comparison of ESRGAN+ vs. WGSR on image 68 from Urban100. (a) Histogram of PSNR differences, (b) Zoomed in HR crop (c) Zoomed in SR crop via ESRGAN+ (d) Zoomed in SR crop via WGSR}
    \label{fig:histogram}
\end{figure}

\noindent \underline{Case 2:} We employ quadrant-based analysis of PSNR differences to compare ESRGAN+ (pixel-loss) vs. WGSR (wavelet-loss) on BSD100 and Urban100. Figure \ref{fig:histogram}(a) shows the histogram of PSNR differences over BSD100 and Urban100, which reveals that while the average PSNR difference shows WGSR outperforms ESRGAN+ by 1.845 dB, there exist significant outlier images, where WGSR surpasses ESRGAN+ by more than 4~dB. Figure \ref{fig:histogram}(b)-(d) depicts one such image (Image 68 from Urban100), where WGSR was able to suppress hallucinations and preserve structured edges more effectively than ESRGAN+, demonstrating the advantages of the including the Wavelet loss term in WGSR for images with sharp transitions. Unfortunately, this disparity in performance is not accurately captured by average PSNR.

\vspace{-8pt}
\begin{minipage}{0.9\columnwidth}
  \begin{algorithm}[H]
  \caption{Top 1\% Error PSNR (PSNR99)}
  \begin{algorithmic}[1]
    \STATE \textbf{Input:} Ground-truth image $\text{HR}$, SR image $\text{SR}$
    \STATE Compute squared error per pixel on Y-channel:
          \vspace{-1em}
      \[
      E \leftarrow (HR - SR)^2
      \]
      \vspace{-2em}
    \STATE Rank pixel-wise errors and select the highest 1\%:
              \vspace{-24pt}

      \[
      E_{\text{top}} \leftarrow \text{R1\%}(E)
      \]
            \vspace{-16pt}

    \STATE Compute the mean of selected errors:
                \vspace{-1em}

      \[
      \text{MSE}_{\text{top}} \leftarrow M(E_{\text{top}}) = \frac{1}{K}\sum E_{\text{top}}
      \]
                  \vspace{-1em}

      where \(K\) is the number of pixels in the top 1\%.
    \STATE Compute 
                \vspace{-2em}
      \[
      \text{PSNR99} \leftarrow 20 \log_{10}\!\Biggl(\frac{255}{\sqrt{\text{MSE}_{\text{top}}}}\Biggr)
      \]
                  \vspace{-15pt}

    \STATE \textbf{Return:} PSNR99
  \end{algorithmic}
  \label{algorithm1}
  \end{algorithm}
\end{minipage}
\vspace{5pt}

\noindent \underline{Case 3:} Finally, we compare the proposed quadrant-based analysis to compare a GAN-based model (WGSR) with a diffusion model for images with different characteristics. 

Table \ref{tab:diff_analysis} further reveals how different architectures, such as diffusion models and GANs 
affect model performance for various image characteristics. For instance, the WGSR-ResShift comparison demonstrates that WGSR outperforms ResShift on easy-edge images, which are highly prone to hallucinations. This issue is mitigated by incorporating wavelet loss, revealing a key insight: tailoring model design and loss functions can lead to significant performance improvements for different image characteristics, insights that would have remained hidden with average metrics.

\begin{table}[t!]
\centering
\resizebox{\columnwidth}{!}{
\begin{tabular}{lccccc}
\toprule
\multicolumn{1}{c}{} & \multicolumn{2}{c}{\textbf{Easy}} & \multicolumn{2}{c}{\textbf{Hard}} & \multirow{2}{*}{\makecell{\textbf{Global}\\\textbf{Average}}} \\
\cmidrule(lr){2-3}\cmidrule(lr){4-5}
\multicolumn{1}{c}{} & \textbf{Texture} & \textbf{Edge} & \textbf{Texture} & \textbf{Edge} &  \\
\midrule
\multicolumn{6}{l}{\textbf{WGSR~\cite{WGSR} vs. ESRGAN+~\cite{ESRGAN_plus} }} \\
PSNR     & 1.904   & 1.823   & 1.776   & 1.868   & 1.845   \\
PSNR99   & 1.625   & 1.750   & 1.529   & 1.628   & 1.633 \\
CLIPIQA \cite{CLIPIQA}  & -0.083   & -0.063   & -0.024   & -0.017   & -0.047   \\
\midrule 
\multicolumn{6}{l}{\textbf{WGSR~\cite{WGSR} vs. ResShift~\cite{ResShift}}} \\
PSNR     & 0.918   & 1.190   & 0.961   & 0.989   & 1.012 \\
PSNR99   & 0.966   & 1.515   & 1.015   & 1.003   & 1.119   \\
CLIPIQA \cite{CLIPIQA} & -0.005   & 0.015   & 0.022   & 0.020   & 0.013   \\
\bottomrule
\end{tabular} \vspace{-10pt}
}
\caption{Quadrant-based analysis of PSNR differences between SR models on Urban100 \cite{urban100_cite} and BSD100 \cite{bsd100_cite}.}
\label{tab:diff_analysis}
\end{table}

These findings in the three cases confirm that averaging PSNR over an entire dataset obscures the strengths and weaknesses of models on specific image characteristics. Moreover, partitioning based on HFI confirms our hypothesis that HFI correlates strongly with PSNR. For finer granularity, the HFI and RIEI axes can be subdivided into three or more levels, creating more subclasses.

\vspace{-4pt}
\section{Conclusion}
\label{sec:conclusion}
\vspace{-12pt}

Reporting average measures (such as PSNR) over a dataset often misses the nuances of model performance, obscuring the relative strengths and weaknesses of models.
Since evaluating SR models on image-basis is not practical, we propose quadrant-based analysis to show the strengths and weaknesses of models/loss functions on images classified as easy/difficult and edge/texture.

As for future research, the proposed RIEI and HFI measures can be used to create mixture of experts SR models, where an input image may be delegated to the appropriate expert model fine-tuned on images with similar RIEI and HFI characteristics at test-time.
Furthermore, PSNR99 can be used as a loss term to minimize the maximum errors generated by SR models whereas RIEI measure can be used as a loss term to minimize artifacts. 
\vspace{-22pt}

\bibliographystyle{IEEEbib}
\bibliography{strings,refs}

\begin{thebibliography}{10}

\bibitem{Agustsson_2017_CVPR_Workshops}
E~Agustsson and R.~Timofte,
\newblock ``{NTIRE 2017 Challenge} on single image super-resolution: Dataset and study,''
\newblock in {\em IEEE/CVF Conf. on Comp. Vision and Patt. Recog. Workshops (CVPRW)}, 07 2017.

\bibitem{cite_LSDIR}
Y.~Li et~al.,
\newblock ``{LSDIR: A} large scale dataset for image restoration,''
\newblock in {\em IEEE/CVF Conf. Comp. Vis. Patt. Recog. Workshop (CVPRW)}, 2023, pp. 1775--1787.

\bibitem{tong_densenet}
T.~Tong, G.~Li, X.~Liu, and Q.~Gao,
\newblock ``Image super-resolution using dense skip connections,''
\newblock in {\em IEEE Int. Conf. Comp. Vis. (ICCV)}, 2017, pp. 4809--4817.

\bibitem{zhang_res_dense}
Y.~Zhang, Y.~Tian, Yu~Kong, B.~Zhong, and Y.~Fu,
\newblock ``Residual dense network for image super-resolution,''
\newblock in {\em IEEE Conf. on Comp. Vis. Patt. Recog. (CVPR)}, 2018, pp. 2472--2481.

\bibitem{zhang2018rcan}
Y.~Zhang, K.~Li, Kai Li, L.~Wang, B.~Zhong, and Yun Fu,
\newblock ``Image super-resolution using very deep residual channel attention networks,''
\newblock in {\em European Conf. Comp. Vision (ECCV)}, 2018.

\bibitem{niu_han}
B.~Niu, W.~Wen, W.~Ren, X.~Zhang, L.~Yang, S.~Wang, K.~Zhang, X.~Cao, and H.~Shen,
\newblock ``Single image super-resolution via a holistic attention network,''
\newblock in {\em Euro. Conf. Comp Vis. (ECCV)}, 2020, p. 191–207.

\bibitem{Liang2021SwinIRIR}
J.~Liang, J.~Cao, G.~Sun, K.~Zhang, L.~Van Gool, and R.~Timofte,
\newblock ``{SwinIR: I}mage restoration using swin transformer,''
\newblock {\em IEEE/CVF Int. Conf. on Comp. Vision (ICCV) Workshops}, pp. 1833--1844, 2021.

\bibitem{chen2023activating}
X.~Chen, X.~Wang, J.~Zhou, Yu~Qiao, and C.~Dong,
\newblock ``Activating more pixels in image super-resolution transformer,''
\newblock in {\em IEEE/CVF Conf. on Comp. Vis. Patt. Recog. (CVPR)}, June 2023, pp. 22367--22377.

\bibitem{ESRGAN_plus}
N.~C. Rakotonirina and A.~Rasoanaivo,
\newblock ``{ESRGAN+: F}urther improving enhanced super-resolution generative adversarial network,''
\newblock in {\em IEEE ICASSP}, May 2020, p. 3637–3641.

\bibitem{details_or_artifacts}
J.~Liang, H.~Zeng, and L.~Zhang,
\newblock ``Details or artifacts: A locally discriminative learning approach to realistic image super-resolution,''
\newblock in {\em IEEE/CVF Conf. on Comp. Vision and Patt. Recog. (CVPR)}, 2022, pp. 5657--5666.

\bibitem{srooe_Park_2023_CVPR}
S.~H. Park, Y.~S. Moon, and N.~Ik Cho,
\newblock ``Perception-oriented single image super-resolution using optimal objective estimation,''
\newblock in {\em IEEE/CVF Conf. Comp. Vis. Pat. Rec. (CVPR)}, 2023, pp. 1725--1735.

\bibitem{freq_sep}
M.~Fritsche, S.~Gu, and R.~Timofte,
\newblock ``Frequency separation for real-world super-resolution,''
\newblock in {\em IEEE/CVF Int. Conf. on Comp. Vision Workshop (ICCVW)}, 2019, pp. 3599--3608.

\bibitem{GuidedFreqSep}
Y.~Zhou, W.~Deng, T.~Tong, and Q.~Gao,
\newblock ``Guided frequency separation network for real-world super-resolution,''
\newblock in {\em IEEE Conf. Comp. Vis. Patt. Recog. Workshops (CVPRW)}, 2020, pp. 1722--1731.

\bibitem{PDASR}
Y.~Zhang, Bo~Ji, J.~Hao, and A.~Yao,
\newblock ``Perception-distortion balanced {ADMM} optimization for single-image super-resolution,''
\newblock in {\em European Conf. on Comp. Vision (ECCV)}, 2022.

\bibitem{WGSR}
C.~Korkmaz, A.~M. Tekalp, and Z.~Dogan,
\newblock ``Training generative image super-resolution models by wavelet-domain losses enables better control of artifacts,''
\newblock in {\em IEEE/CVF Conf. on Computer Vision and Patt. Recog. (CVPR)}, 2024.

\bibitem{cheng2023newsuperresolutionmeasurementperceptual}
Sheng Cheng,
\newblock ``A new super-resolution measurement of perceptual quality and fidelity,'' 2023.

\bibitem{srqc}
Greeshma~M S and Bindu V~R,
\newblock ``Super-resolution quality criterion (srqc): a super-resolution image quality assessment metric,''
\newblock {\em Multimedia Tools and Applications}, vol. 79, pp. 1--22, 12 2020.

\bibitem{Wu_2024_CVPR}
R.~Wu, T.~Yang, L.~Sun, Z.~Zhang, S.~Li, and L.~Zhang,
\newblock ``{SeeSR:} towards semantics-aware real-world image super-resolution,''
\newblock in {\em IEEE/CVF Conf. on Comp. Vision and Patt. Recog. (CVPR)}, June 2024, pp. 25456--25467.

\bibitem{lpips}
R.~Zhang, P.~Isola, A.~A. Efros, E.~Shechtman, and O.~Wang,
\newblock ``The unreasonable effectiveness of deep features as a perceptual metric,''
\newblock in {\em IEEE/CVF Conf. on Comp. Vision and Patt. Recog. (CVPR)}, 2018, pp. 586--595.

\bibitem{niqe}
A.~Mittal, R.~Soundararajan, and A.~C. Bovik,
\newblock ``Making a “completely blind” image quality analyzer,''
\newblock {\em IEEE Signal Proc. Letters}, vol. 20, no. 3, pp. 209--212, 2013.

\bibitem{9897278}
C.~Korkmaz, A.~M. Tekalp, and Z.~Doğan,
\newblock ``{MMSR: M}ultiple-model learned image super-resolution benefiting from class-specific image priors,''
\newblock in {\em IEEE Int Conf. on Image Proc. (ICIP)}, 2022, pp. 2816--2820.

\bibitem{bsd100_cite}
D.~Martin, C.~Fowlkes, D.~Tal, and J.~Malik,
\newblock ``A database of human segmented natural images and its application to evaluating segmentation algorithms and measuring ecological statistics,''
\newblock in {\em Int. Conf. Comp. Vis. (ICCV)}, 2001, pp. 416--423.

\bibitem{urban100_cite}
Jia-Bin Huang, Abhishek Singh, and Narendra Ahuja,
\newblock ``Single image super-resolution from transformed self-exemplars,''
\newblock in {\em IEEE Conf. on Comp. Vis. Patt. Recog. (CVPR)}, 2015, pp. 5197--5206.

\bibitem{CLIPIQA}
J.~Wang, K.~CK Chan, and C.~C. Loy,
\newblock ``Exploring {CLIP} for assessing the look and feel of images,''
\newblock in {\em Assoc. Adv. Artif. Intel. (AAAI)}, 2023.

\bibitem{ResShift}
Z.~Yue, J.~Wang, and C.~Change Loy,
\newblock ``Efficient diffusion model for image restoration by residual shifting,''
\newblock {\em IEEE Trans. on Patt. Anal. and Mach. Intel.}, vol. 47, no. 1, 2025.

\end{thebibliography}

\end{document}